\documentclass[letterpaper, 10 pt, journal, twoside]{IEEEtran}

\usepackage{hyperref}

  \usepackage{graphics}           
\usepackage{times}              
\usepackage{amsmath}            
\usepackage{amssymb}            
\usepackage{graphicx}
\usepackage{algorithm}
\usepackage[noend]{algpseudocode}
\usepackage{booktabs}
\usepackage{color}
\definecolor{instructioncolor}{rgb}{.5,.5,.5}

\usepackage[font=small]{caption}

\def\secref#1{Sec.~\ref{#1}}
\def\figref#1{Fig.~\ref{#1}}
\def\tabref#1{Tab.~\ref{#1}}
\def\eqref#1{Eq.~(\ref{#1})}

\makeatletter
\usepackage{xspace}
\DeclareRobustCommand\onedot{\futurelet\@let@token\@onedot}
\def\@onedot{\ifx\@let@token.\else.\null\fi\xspace}

\def\etal{{et al}\onedot}
\makeatother

\def\etalcite#1{\etal~\cite{#1}}

\usepackage{array}
\newcolumntype{L}[1]{>{\raggedright\let\newline\\\arraybackslash\hspace{0pt}}m{#1}}
\newcolumntype{C}[1]{>{\centering\let\newline\\\arraybackslash\hspace{0pt}}m{#1}}
\newcolumntype{R}[1]{>{\raggedleft\let\newline\\\arraybackslash\hspace{0pt}}m{#1}}

\renewcommand{\b}[1]{\mbox{\boldmath$#1$}}

\renewcommand{\v}[1]{{\b #1}}

\graphicspath{{pics/}}

\usepackage{todonotes}
\usepackage{booktabs}
\usepackage{multirow}
\usepackage{bm}
\usepackage[normalem]{ulem}

\usepackage{pifont}

\usepackage{transparent}
\usepackage[acronyms, shortcuts]{glossaries}
\glsdisablehyper
\newacronym{cnn}{CNN}{convolutional neural network}
\newacronym{rnn}{CNN}{recurrent neural network}
\newacronym{mos}{MOS}{moving object segmentation}
\newacronym{bev}{BEV}{bird's eye view}
\newacronym{knn}{kNN}{k-nearest neighbor}
\newacronym{fifo}{FIFO}{first in, first out}
\newacronym{iou}{IoU}{intersection-over-union}
\newacronym{icp}{ICP}{iterative closest point}
\newacronym{slam}{SLAM}{simultaneous localization and mapping}
\newacronym{tsdf}{TSDF}{truncated signed distance function}

\newcommand{\fdmos}{4DMOS\xspace}
\newcommand{\kiss}{KISS-ICP\xspace}
\newcommand{\lmnet}{LMNet\xspace}
\newcommand{\rvmos}{RVMOS\xspace}
\newcommand{\motionseg}{MotionSeg3D\xspace}
\newcommand{\vdbfusion}{VDBFusion\xspace}

\newcommand{\semkitti}{SemanticKITTI\xspace}
\newcommand{\kittitracking}{KITTI Tracking\xspace}
\newcommand{\apollo}{Apollo\xspace}
\newcommand{\apollofull}{Apollo Columbia Park MapData\xspace}
\newcommand{\nuscenes}{nuScenes\xspace}
\newcommand{\mosbench}{SemanticKITTI MOS benchmark\xspace}

\newcommand{\map}{local map\xspace}
\newcommand{\Map}{Local Map\xspace}
\newcommand{\Belief}{Volumetric Belief\xspace}
\newcommand{\belief}{volumetric belief\xspace}
\newcommand{\bayesupdate}{l\left(m_i \mid \mathcal{P}_{t}, \mathcal{S}_{t}\right)\xspace}
\newcommand{\bayesrecursive}{l\left(m_i \mid \mathcal{P}_{1:t-1}, \mathcal{S}_{1:t-1}\right)\xspace}
\newcommand{\bayesprior}{l(m_i)\xspace}
\newcommand{\bayesbelief}{l\left(m_i \mid \mathcal{P}_{1:t}, \mathcal{S}_{1:t}\right)\xspace}

\newcommand{\oursscan}{Scan}
\newcommand{\oursbelief}{\Belief, Scan Only}
\newcommand{\oursmap}{\Belief, No Delay}
\newcommand{\oursdelay}{\Belief}

\newcommand{\state}{dynamic occupancy\xspace}
\newcommand{\outputs}{logits\xspace}

\newcommand{\bayesfilter}{binary Bayes filter\xspace}

\newcommand{\strategy}{receding horizon strategy\xspace}

\usepackage{flushend}

\title{Building Volumetric Beliefs \\ for Dynamic Environments Exploiting \\ Map-Based Moving Object Segmentation}

\author{Benedikt Mersch, Tiziano Guadagnino, Xieyuanli Chen, Ignacio Vizzo, Jens Behley, and Cyrill Stachniss%
	\thanks{Manuscript received: March 10, 2023; Revised: June 06, 2023; Accepted: June 28, 2023. This paper was recommended for publication by Editor Javier Civera upon evaluation of the Associate Editor and Reviewers' comments.}%
	\thanks{This work has partially been funded by the European Union’s Horizon 2020 research and innovation programme under grant agreement No~101017008~(Harmony) and by the European Union’s Horizon Europe research and innovation programme under grant agreement No~101070405~(DigiForest).}%
	\thanks{All authors are with the University of Bonn, Germany. Cyrill Stachniss is additionally with the Department of Engineering Science at the University of Oxford, UK, and with the Lamarr Institute for Machine Learning and Artificial Intelligence, Germany.}%
	\thanks{Digital Object Identifier (DOI): see top of this page.}
}

\begin{document}
\maketitle

\markboth{IEEE Robotics and Automation Letters. Preprint Version. Accepted June, 2023}
{Mersch \MakeLowercase{\textit{et al.}}: Building Volumetric Beliefs for Dynamic Environments Exploiting Map-Based Moving Object Segmentation}

\begin{abstract}
	Mobile robots that navigate in unknown environments need to be constantly aware of the dynamic objects in their surroundings for mapping, localization, and planning. It is key to reason about moving objects in the current observation and at the same time to also update the internal model of the static world to ensure safety. In this paper, we address the problem of jointly estimating moving objects in the current 3D LiDAR scan and a \map of the environment. We use sparse 4D convolutions to extract spatio-temporal features from scan and \map and segment all 3D points into moving and non-moving ones. Additionally, we propose to fuse these predictions in a probabilistic representation of the dynamic environment using a Bayes filter. This \belief models, which parts of the environment can be occupied by moving objects. Our experiments show that our approach outperforms existing moving object segmentation baselines and even generalizes to different types of LiDAR sensors. We demonstrate that our \belief fusion can increase the precision and recall of moving object segmentation and even retrieve previously missed moving objects in an online mapping scenario.
\end{abstract}

\section{Introduction}
\label{sec:intro}
\IEEEPARstart{S}{egmenting} moving and non-moving objects is key for mobile robots operating in dynamic environments. It is an important step for online applications like mapping~\cite{ruchti2018icra,vizzo2022sensors}, localization~\cite{henein2018arxiv,pfreundschuh2021icra}, planning~\cite{kuemmerle2013icra}, or occupancy prediction~\cite{hoermann2018icra}. To solve such tasks, a robot needs to reason about which parts of the environment are moving and which are not in an online fashion. For successful navigation and planning, this knowledge should not only be limited to what the robot currently perceives but rather be integrated into a representation of the environment.

\begin{figure}[t]
	\centering
	\def\svgwidth{0.99\linewidth}
	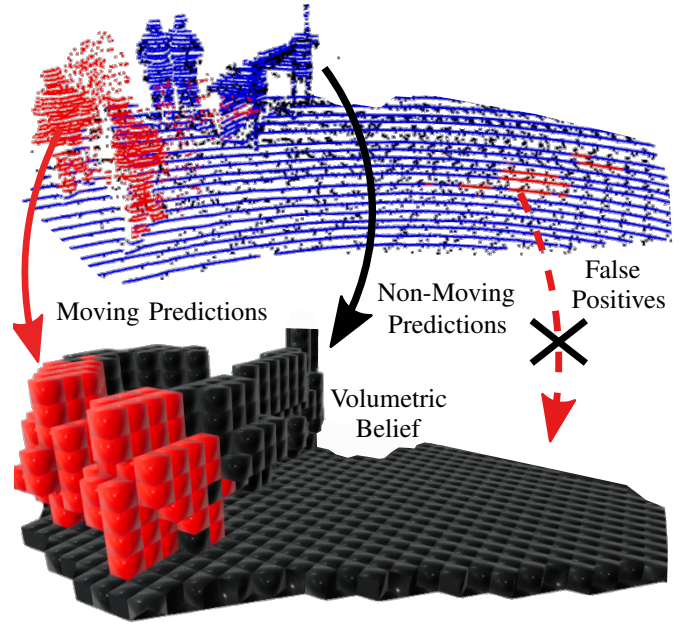
	\caption{Our approach identifies moving objects \textit{(red)} in the current scan \textit{(blue)} and the \map \textit{(black)} of the environment. We maintain a \belief map representing the dynamic environment and fuse new predictions in a probabilistic fashion. This allows us to reject false positive predictions that contradict our \belief.}
	\label{fig:motivation}
\end{figure}

In this paper, we investigate the problem of segmenting moving objects in both current and past 3D LiDAR scans. Additionally, we maintain a 3D model of the environment representing our belief about which part of the space can contain moving objects as depicted in~\figref{fig:motivation}. We update the belief online by fusing our predictions in a probabilistic manner to increase precision and recall of~\ac*{mos}. Also, for~\ac*{slam}, it is of central interest to estimate, which parts of the environment are dynamic. The knowledge of moving objects can be directly integrated into the optimization as an object motion estimation as shown by Henein~\etalcite{henein2018arxiv}, in this example done for rigid body motion. Pfreundschuh~\etalcite{pfreundschuh2021icra} and Chen~\etalcite{chen2021ral,chen2019iros} demonstrate the effectiveness of moving object segmentation for data associations. An alternative strategy for localization and long-term planning is to build a map and clean it from traces of dynamic objects in a post-processing step,~\cite{arora2021ecmr,arora2023jras,kim2020iros,lim2021ral}. Thus, the addressed estimation problem has multiple relevant applications in robotics.

If building a static map is required online, one way is to segment each incoming scan into moving and non-moving and then integrate only the static points into the map~\cite{chen2021ral}. In this setup, each segmentation is done independently of previous predictions. The downside of this approach is that it is not straightforward to recover a missed moving object that was added to the map. Recently, \fdmos~\cite{mersch2022ral} improved the segmentation robustness by re-estimating moving objects in a scan after receiving more observations and fusing them in a \bayesfilter. However, the \ac*{mos} robustness can only be improved as long as the corresponding scan is within the limited buffer of past scans that \fdmos and related approaches consider for prediction. Also, the idea of using a local buffer assumes that the movement of an object can be identified from consecutive measurements. This usually holds for most rotating LiDAR scanners that scan the surroundings with a regular scanning pattern at high frequency, but not for scanners with a limited field of view or irregular sampling patterns~\cite{lin2019iros-larl}.

The main contributions of this paper are two-fold. First, we propose an approach to predict moving objects in a \map constructed using all past LiDARs measurements recorded in this area without limiting the time horizon. Second, we build and maintain a \belief map and fuse new predictions in a voxel-wise \bayesfilter to previous estimates online, which increases robustness and corrects previously wrong predictions. In sum, we make four key claims:
Our approach is able to
(i) accurately segment an incoming LiDAR scan into moving and non-moving objects based on a \map of past observations,
(ii) generalize well to new environments and sensor setups while achieving state-of-the-art performance,
(iii) increase the precision and recall of~\acl*{mos} by fusing multiple predictions into a \belief,
(iv) recover from wrong predictions for online mapping through a \belief.
These claims are backed up by the paper and our experimental evaluation. Our code, pre-trained models, and labels for evaluation are available at \mbox{\url{https://github.com/PRBonn/MapMOS}}.

\section{Related Work}
\label{sec:related}
\textbf{Online LiDAR~\ac*{mos}} is usually achieved by comparing the current scan against the past, with the goal of segmenting the corresponding point cloud into moving and non-moving parts~\cite{chen2022ral,mersch2022ral}. Yoon~\etalcite{yoon2019crv} identify moving objects based on the residual between two scans, free space filtering, and region growing post-processing. One drawback of such an approach is that objects can be temporarily occluded, which makes it hard to identify motion only considering two scans.

Subsequent works extend the temporal horizon of past information that is used for prediction. Processing more points at full resolution is often computationally demanding. Therefore, most methods project the data into a lower-dimensional representation~\cite{chen2021ral,kim2022ral,sun2022iros}. Chen~\etalcite{chen2021ral} use past residual images together with a semantic segmentation network to segment a range-image representation of the scan into moving and non-moving. Work by Kim~\etalcite{kim2022ral} extends this idea by additionally predicting movable and non-movable objects from semantics. Sun~\etalcite{sun2022iros} propose a point refinement module to reduce the effect of imprecise boundaries, sometimes referred to as ``label bleeding'' for range image-based segmentation~\cite{milioto2019iros}.

The problem of label bleeding is also addressed in~\fdmos~\cite{mersch2022ral} by predicting moving objects in the voxelized 4D space without prior projection. It assumes that the motion of an object is visible within a limited time horizon of consecutive past scans, which are aggregated to a sparse 4D point cloud. By shifting this temporal window, the prediction of previous scans can be refined by fusing them in a point-wise \bayesfilter. To reduce the effort of labeling, Kreutz~\etalcite{kreutz2023wacv} proposed a feature encoding and clustering approach based on a 4D occupancy time series.

Similar to \fdmos, we extract spatio-temporal features using 4D convolution instead of projecting the data. In contrast to the aforementioned methods~\cite{chen2021ral,kim2022ral,mersch2022ral,sun2022iros}, our proposed approach predicts moving objects using the current scan and a voxelized local point cloud of all past scans without limiting the time horizon.

\textbf{Static Map Building --}
A standard approach to obtain a static model of the environment is to only integrate static points based on a scan-wise~\ac*{mos}~\cite{chen2021ral}. Other researchers focused on geometric approaches to obtain a static representation of the environment. For example, occupancy maps divide the space into occupied, free, and unobserved areas~\cite{thrun2005probrobbook}. The static belief of voxels is updated by ray-tracing and recursive Bayesian estimation using an inverse sensor model~\cite{thrun2005probrobbook}. The final map can be used to decide if a new measurement belongs to a dynamic object or not~\cite{wellhausen2017ssrr}. Stachniss and Burgard~\cite{stachniss2005aaai} propose an approach for 2D grid-based localization in non-static environments by clustering possible configurations of the changing environment which improves localization. To cover the full spectrum of temporal changes in the environment, Biber and Duckett~\cite{biber2005rss} update a map based on different time scales.

In contrast, Nuss~\etalcite{nuss2018ijrr} propose to use random finite sets to explicitly model the dynamic state of each grid cell, which has been further used for tasks like occupancy prediction~\cite{hoermann2018icra}. To deal with 3D LiDAR data, Wurm~\etalcite{wurm2010icraws} and Hornung~\etalcite{hornung2013ar} introduced OctoMap which extends occupancy grid mapping to the 3D space by using an octree data structure. Ray-tracing on volumetric occupancy grids has also been researched to remove dynamic objects from a set of LiDAR scans~\cite{gehrung2017isprsannals,schauer2018ral}. Similarly, Pagad~\etalcite{pagad2020icra} use an octree to build an occupancy grid map by first detecting ground and object points and using ray-tracing to update voxel occupancy. Arora~\etalcite{arora2021ecmr,arora2023jras} exploit OctoMap for static map cleaning and leverage ground-segmentation and a voting scheme to deal with unknown points.

Visibility-based methods~\cite{kim2020iros,lim2021ral,pomerleau2014icra} alleviate the computational cost of ray-tracing by checking the consistency of a query point with respect to a pre-built map. For example, Lim~\etalcite{lim2021ral} identify temporarily occluded regions in an accumulated point cloud map based on height discrepancy between query and map. Instead of removing dynamic points, Huang~\etalcite{huang2022eccv} explicitly target the reconstruction of moving objects for 3D scene analysis. To deal with the sparse measurements from moving objects, the authors register multiple point clouds and estimate offset vectors of previously classified moving points.

In our work, we aim at closing the gap between scan-wise online~\ac*{mos} and an offline volumetric representation of the dynamic environment. We propose an approach that segments the current scan as well as previously received measurements into moving and non-moving points and fuses these predictions in a 3D volumetric representation. In contrast to most of the aforementioned approaches, we maintain this belief online and use it to robustify the current prediction and to retrieve previously missed moving objects for online mapping.

\section{Our Approach}
\label{sec:main}
We propose to segment moving objects based on the discrepancy between the current LiDAR frame and a \map consisting of the previously measured scans in that area. Given the current LiDAR frame at time~$t$, we first register it to our current \map as explained in~\secref{sec:registration}. Next, we jointly predict moving objects in the aligned scan and the \map, see~\secref{sec:scan2map}. After that, we fuse these predictions into a probabilistic \belief to maintain a representation of the dynamic environment, see~\secref{sec:belief_update}. We can query the \belief for a set of points as explained in~\secref{sec:belief_query} to obtain the current belief if these points belong to moving objects or not.

\subsection{Scan Registration with \kiss}
\label{sec:registration}
Our approach does not require ground truth poses, it only relies on sequential 3D LiDAR data. When a new measurement is available, we register the scan using \kiss~\cite{vizzo2023ral}, which is a robust odometry pipeline that generalizes well to varying motion profiles and sensor platforms without the need for changing parameters.

Our \map used within this paper is a sparse voxel grid as the one of \kiss. We maintain the original coordinates of the points in the voxels to avoid discretization errors. In contrast to the original \kiss implementation, we additionally store for each point the timestamp of the scan it stems from to maintain temporal information in the \map. Our method directly uses this temporal information to predict moving objects for both the registered scan and the \map.

\subsection{Map-based Moving Object Segmentation}
\label{sec:scan2map}
We start by explaining how to jointly predict moving objects after registering a new LiDAR frame. We exploit two different mechanisms to segment moving objects. First, we consider the spatial discrepancy between the current scan and the \map. This information indicates if an object may have moved with respect to all previous measurements in that area. Second, we identify the motion of objects based on the evolution of the timestamps given by the feature attached to each point. This allows us to segment both the current scan and the \map into moving and non-moving parts.

In contrast to previous works~\cite{chen2021ral,kim2022ral,mersch2022ral,sun2022iros}, our method is not restricted to a fixed set of past scans. This is advantageous in cases a moving object is not fully visible within a short time horizon due to occlusion, limited field of view, or an irregular shooting pattern of the LiDAR\@. In practice, this makes a substantial difference.

\begin{figure}[t]
	\centering
	\def\svgwidth{0.99\linewidth}
	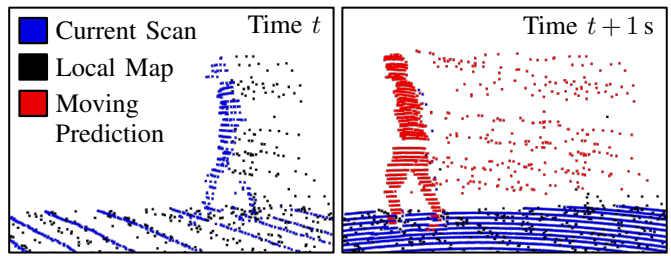
	\caption{Current Scan and \map for two different times with our moving predictions in red. Although our method initially failed to fully identify the moving pedestrian in the beginning \textit{(left)}, we successfully predict it at a later point in time and our method backtraces the corresponding points in the \map \textit{(right)}.}
	\label{fig:backtrace}
\end{figure}

Additionally, instead of predicting moving objects in the current scan or a limited buffer of scans, we segment both the current scan and the \map. Segmenting the \map enables us to identify traces of moving objects that were not segmented in previous scan predictions. This backtracing of dynamic objects allows us to correct initial false negative predictions as shown in~\figref{fig:backtrace}.

Our \map is the voxel grid structure of \kiss, but we store for every point its 4D coordinate (position plus time). To maintain the ordering of scan and \map during the convolutions, we organize them in a 4D tensor. We use the timestamps as features for the points and normalize them based on the minimum and maximum values since we are only interested in their relative difference. This avoids the model overfitting to the sequence lengths and, therefore, the maximum timestamps it has seen during training.

\begin{figure*}[t]
	\centering
	\def\svgwidth{0.99\linewidth}
	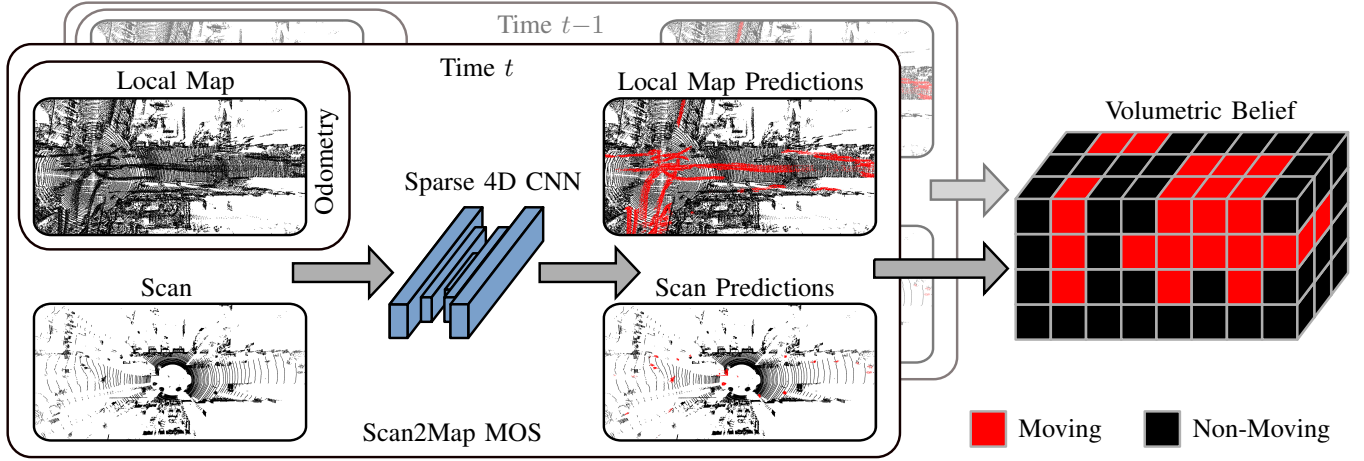
	\caption{Overview our proposed Scan2Map~\acl*{mos} approach and \belief fusion. At time~$t$, we predict moving objects in the current scan and \map using sparse 4D convolutions. Next, we update our \belief about which parts of the environment can contain moving objects based on the previous \belief at time~$t\,{-}\,1$ and our new predictions.}
	\label{fig:overview}
\end{figure*}

At time~$t$, we voxelize the 4D point cloud~$\mathcal{P}_t$ of scan and \map and represent it as a sparse 4D tensor using the MinkowskiEngine~\cite{choy2019cvpr}. Sparse tensors are a more memory-efficient representation for 4D tensor data and allow to directly apply sparse convolutions. We jointly extract spatial and temporal features with sparse 4D convolutions. Our network architecture is a 4D MinkUNet~\cite{choy2019cvpr} with~$1.8$\,Mio parameters. This network first downsamples the points and features in an encoder to extract high-level information and then upsamples both to the original resolution in a decoder. Residual blocks and skip connections help to maintain detailed information about the points and their corresponding features. The last layer predicts the \outputs~$\mathcal{S}_t$ of both current scan and \map points being moving.~\figref{fig:overview} depicts an overview of our approach.

\subsection{\Belief Update}
\label{sec:belief_update}
In this section, we present our approach to fuse per-point \ac*{mos} predictions into a probabilistic \belief. Fusing multiple independent predictions over time can filter out prediction errors from the neural network and has been previously explored on the point-level~\cite{mersch2022ral}. Instead of fusing per-points predictions, our goal is to model, which parts of the environment have a higher probability of containing a dynamic object. Notice that in this case, we do not want to just identify current dynamics, but rather determine which portion of the map is traversed by moving objects. We define this property as \emph{\state}.

We assume that the binary state~$m_i\,{\in}\,\{0,1\}$ of \state for a voxel~$v_i$ does not change over time. Intuitively, this means that if a point falls into a voxel previously occupied by dynamics, we assume that this point also belongs to a moving object. On the other hand, if a voxel was occupied by static points, we do not expect to observe a moving object in this volume. Note that this state definition is different from occupancy grid mapping, where the world is assumed to be static and a fixed occupancy probability of a cell is estimated.

At time~$t$, we predict~$N$ \outputs~$\mathcal{S}_t\,{=}\,\{ s_{t,1}, s_{t,2},\dots s_{t,N} \}$ with~$s_{t,j}\,{\in}\,\mathbb{R}$ for~$N$ points~$\mathcal{P}_t\,{=}\,\{ \v{p}_{t,1}, \v{p}_{t,2}, \dots \v{p}_{t,N}\}$ with~$\v{p}_{t,j}\,{\in}\,\mathbb{R}^4$ as described in ~\secref{sec:scan2map}. It is possible to fuse the \outputs for the current scan but also for the \map points. We provide an experiment in~\secref{exp:ablation} to showcase the results for different fusion strategies. Note that our \belief is not restricted to our \outputs, but that predictions from different sources could be integrated. Our goal is to estimate the joint probability distribution of the \belief map state for all voxels~$\mathcal{M}\,{=}\,\{m_i\}$ reading
\begin{equation}\label{eqn:joint_prob}
	p\left(\mathcal{M} \mid \mathcal{P}_{1:t}, \mathcal{S}_{1:t} \right) = \prod_i p\left(m_i \mid \mathcal{P}_{1:t}, \mathcal{S}_{1:t} \right),
\end{equation}
with $\mathcal{P}_{1:t}$ and $\mathcal{S}_{1:t}$ being the sets of previously measured points and predicted \outputs up to time~$t$, respectively.

After applying Bayes' rule to the right-hand side per-voxel probability distribution, we can derive the recursive binary Bayes filter equations according to Thrun~\etalcite{thrun2005probrobbook}. We use the log-odds notation~$l(x){=}\log \frac{p(x)}{1-p(x)}$ resulting in
\begin{equation}\label{eq:log_odds}
	\begin{split}
		\bayesbelief
		& = \bayesrecursive \\
		& \quad + \bayesupdate - \bayesprior,
	\end{split}
\end{equation}
for updating a single voxel cell belief~$\bayesbelief$. Here,~$\bayesrecursive$ is the recursive term currently stored in the voxel, which aggregates the previous predictions,~$\bayesupdate$ is the update term for the voxel which integrates the predictions at the current time~$t$, and~$\bayesprior$ are the log-odds of the prior probability~$p_0$. We do not assume to have prior knowledge about the \state of a voxel~$v_i$ and therefore set it to~$p_0=0.5$.

The remaining step is to get a per-voxel update~$\bayesupdate$ from the points~$\mathcal{P}_t$ and \outputs~$\mathcal{S}_t$. The prediction~$s_{t,j}~{\in}~\mathcal{S}_t$ at time~$t$ for a single point~$\v{p}_{t,j}$ with index~$j$ indicates if it belongs to a moving object or not. Since multiple points with different \outputs can end up in the same voxel, we need to aggregate their information and take the arithmetic mean of \outputs inside a voxel~$i$ resulting in
\begin{equation}\label{eq:update}
	\bayesupdate = \frac{\sum_{j \in \mathcal{V}_{t,i}} s_{t,j}}{\left|\mathcal{V}_{t,i}\right|},
\end{equation}
where $\mathcal{V}_{t,i} = \{ j\mid\v{p}_{t,j} \in v_i\}$ is the set of points falling into the voxel~$v_i$ at time~$t$ and $|\mathcal{V}_{t,i}|$ is the cardinality of the set. Taking the arithmetic mean of per-point log-odds corresponds to the geometric mean of the individual likelihoods of a point being moving. Likelihood aggregation using the geometric mean has been previously used in Monte-Carlo localization sensor model designs~\cite{zimmerman2022iros}.

We implement our \belief as a hash table, which is a more memory-efficient representation compared to dense 3D arrays,~\cite{niessner2013siggraph,vizzo2023ral}. Each 3D voxel~$v_i$ stores the log-odds belief~$\bayesbelief$ about its \state state~$m_i\,{\in}\,\{0,1\}$ after integrating predictions up to time~$t$.

\subsection{\Belief Query}
\label{sec:belief_query}
For a given set of points, we can query our \belief by indexing the corresponding voxels~$v_i$ and converting the log-odds beliefs~$\bayesbelief$ to a posterior probability $p$ using~$p(x)\,{=}\,\frac{e^{l(x)}}{1+e^{l(x)}}$. We assume that a point is moving if the probability is larger than~$0.5$. Note that the voxel size of our \belief needs to be appropriate since the underlying assumption is that all points inside a voxel share the same \state state. This assumption is violated if the voxel size is too large.

\subsection{Online Mapping}
\label{sec:mapping}
For online mapping, we are interested in accurately removing moving points. We experienced discretization effects at the boundaries of moving objects, for example, false negative predictions on the wheels of vehicles close to the ground. To achieve sub-voxel accuracy and a high recall for identifying moving objects, we combine the filtered voxel-wise \belief with the point-wise scan prediction for online mapping. We demonstrate this in an experiment in~\secref{exp:mapping}.

\subsection{Implementation Details}
\label{sec:implementation}
We set the voxel size used for downsampling the scans in our odometry system to~$0.5$\,m. We train our 4D CNN by supervising the prediction for scan and \map points using the cross-entropy loss for~$100$ epochs and save the model performing best on the validation set. Since some sequences of the training set do not contain a lot of moving objects, we skip a batch if the ratio between moving and static points is less than~$0.1$\,\%. Next, we crop a rectangular patch of the scenes and augment the batch by rotating, flipping, and scaling. Lastly, we randomly drop points with a dropout rate sampled from the interval~$[0,0.5]$ to vary the density of the point clouds. One epoch takes less than 25\,min on an NVIDIA RTX A5000.

For choosing a voxel size of our \belief map, one needs to trade off between computational efficiency and accuracy due to the discretization. For our experiments, we set the fixed voxel size to~$0.25$\,m. Additionally, we clip the \belief map at~$150$\,m.

\section{Experimental Evaluation}
\label{sec:exp}
The main focus of this work is an approach to identify moving objects in the current LiDAR frame and a \map of aggregated past scans and to fuse these predictions into a probabilistic \belief map. We present our experiments to show the capabilities of our method. The results of our experiments also support our key claims, which are:
Our approach (i) accurately segments an incoming LiDAR scan into moving and non-moving objects based on a \map of past observations,
(ii) generalizes well to new environments and sensor setups while achieving state-of-the-art performance,
(iii) increases the precision and recall of~\ac*{mos} by fusing multiple predictions into a \belief,
(iv) recovers from wrong predictions for online mapping through a \belief.

\subsection{Datasets, Metrics, and Baselines}
For the following experiments, we train all models on the moving labels of the \semkitti~\cite{behley2021ijrr,behley2019iccv} training sequences 00-07 and 09-10 and use sequence 08 for validation. We do not use the KITTI pose information since our approach registers the scans using \kiss~\cite{vizzo2023ral} as described in~\secref{sec:registration}.

Besides the commonly used \mosbench~\cite{chen2021ral} based on the \semkitti labels, we also evaluate and compare our approach on a labeled sequence from the \kittitracking~\cite{geiger2012cvpr} dataset recorded with the same sensor setup in a street with a lot of moving pedestrians. We additionally report results on a subset of the \apollofull~\cite{lu2019cvpr} with labels provided by Chen~\etalcite{chen2022ral}. This data is recorded with the same sensor, but in a different city environment.

To push the generalization capabilities of~\ac*{mos} approaches, we test the models trained on \semkitti with 64 vertical beams at~$10$\,Hz frequency on the \nuscenes~\cite{caesar2020cvpr} dataset, which has 32 vertical beams at~$20$\,Hz. We evaluate the~\ac*{mos} for \nuscenes based on the moving labels from the annotated keyframes of the~$150$ validation sequences.

We assess the performance using the commonly known~\ac*{iou}~\cite{everingham2010ijcv} of the moving points and additionally precision and recall in~\secref{exp:ablation}.

We compare our method to the projection-based baselines \lmnet~\cite{chen2021ral}, \motionseg~\cite{sun2022iros}, and \rvmos~\cite{kim2022ral}. For \motionseg, we show the results without (v1) and with the proposed point refinement (v2). \fdmos~\cite{mersch2022ral} applies sparse 4D convolutions, but on a limited buffer of aggregated, registered past scans.

\subsection{Moving Object Segmentation Performance}
\label{exp:kitti}

\begin{table}[t]
	\centering
	\begin{tabular}{lcc}
		\toprule
		Method                               & Test 11-21        & Validation 08 \\
		\midrule
		\lmnet~\cite{chen2021ral}            & 58.3              & 66.4          \\
		\motionseg, v1~\cite{sun2022iros}    & 62.5              & 68.1          \\
		\motionseg, v2~\cite{sun2022iros}    & 64.9              & 71.4          \\
		\fdmos, delayed~\cite{mersch2022ral} & 65.2              & 77.2          \\
		Ours, \oursscan                      & 65.9              & 83.8          \\
		Ours, \oursdelay                     & \textbf{66.0}     & \textbf{86.1} \\
		\midrule
		\rvmos~\cite{kim2022ral}$^*$         & \textbf{73.3}$^*$ & 71.2$^*$      \\
		\bottomrule
	\end{tabular}
	\caption{Comparison of average moving~\ac*{iou} on the \semkitti validation sequence 08 and the \mosbench~\cite{chen2021ral}. Best results in bold. The\,{$^*$}\,indicates that the approach additionally \emph{exploits semantic labels}.}
	\label{tab:kitti}
\end{table}

In the first experiment, we evaluate how well our approach segments a scan into moving and non-moving points by using a \map of past observations. We show the originally reported baseline results on the \semkitti validation set and the \mosbench. To provide fair comparisons, we only consider approaches, which are trained and validated on the original \semkitti split. This eliminates the positive bias of using additional training data~\cite{chen2022ral,sun2022iros}.

We evaluate both the predictions of the current scan (referred to as ``\oursscan'') and the \belief with a delay of 10 scans (referred to as ``\oursdelay''). The choice of 10 scans is an initial estimate that trades off the ability to correct previous wrong estimates and the required waiting time. Besides fusing all scan predictions, we decide to only integrate the \map points that we predict to be moving. This has two reasons: First, we are mainly interested in the moving objects in the \map that we have missed in previous scan predictions. Second, integrating all \map points reduces the runtime of the system. The delay of 10 scans helps to get a more informed belief about the voxels with additional \map predictions before querying their state.

One can see in~\tabref{tab:kitti} that our \belief helps to improve the results on the validation sequence, whereas the effect is smaller on the test set. We further investigate the effect of the \belief in~\secref{exp:ablation}. Our approach outperforms \fdmos, showing that not limiting past information is beneficial for~\ac*{mos}. In general, we rank second best on the hidden test set and are only outperformed by \rvmos, which requires additional semantic labels for training, while all other approaches just use the moving object labels. Our approach using the \belief achieves the highest result on the validation set with~$86.1$\,\%~\ac*{iou} for the moving points.

Our~\ac*{mos} model runs at 12\,Hz for the \mosbench using an NVIDIA RTX A5000. We implemented the \belief update and querying in C++ and it runs at 44\,Hz on a Intel(R) Xeon(R) W-1290P CPU @ 3.70\,GHz processor with multi-threading.

\subsection{Generalization Capabilities}
\label{exp:generalization}
The next experiment analyzes how well our approach generalizes to new environments and sensor setups. Since~\ac*{mos} is often a supervised task and labeling is expensive, generalization is an important property. We provide an experiment in~\tabref{tab:generalization} that realizes different levels of domain shift and compare how well the approaches generalize.

All baselines require external pose information, which we acquire using \kiss for a fair comparison. Note that in the case of \fdmos, we also report the result of segmenting the most recent scan to compare the online performance before refining with the originally proposed \strategy and \bayesfilter. Unfortunately, the code for \rvmos is not publicly available so we cannot run it on these additional datasets.

One can see that the projection-based approaches \lmnet and \motionseg perform worse on the highly crowded \kittitracking sequence 19. Their performance drops even further on the \apollo dataset. We believe this is because they implicitly overfit to the calibration of the LiDAR sensor, such as mounting location and intensity measurements.

In contrast, \fdmos and our approach only use the temporal information of the scans and therefore generalize well to a new sensor calibration. We again obtain the best result using our \belief with a delay of 10 scans (referred to as ``\oursdelay'') and outperform \fdmos.

For the \nuscenes dataset, we cannot evaluate the pre-trained models for the projection-based approaches in a fair comparison, because the range image dimensions change due to the different vertical resolutions of the sensors. Both \fdmos and our approach are still able to segment moving objects, but the average moving~\ac*{iou} is lower. Here, the strategy of \fdmos shows the best results. When comparing the current scan predictions only, we again achieve a better result in terms of moving~\ac*{iou}.

\begin{table}[]
	\centering
	\begin{tabular}{lccc}
		\toprule
		                                     & KITTI~\cite{geiger2012cvpr} & \apollo~\cite{lu2019cvpr} & \nuscenes~\cite{caesar2020cvpr} \\
		Method                               & Tracking 19                 &                           & Validation                      \\
		\midrule
		\lmnet~\cite{chen2021ral}            & 45.3                        & 13.7                      & n/a                             \\
		\motionseg, v1~\cite{sun2022iros}    & 54,6                        & 6.5                       & n/a                             \\
		\motionseg, v2~\cite{sun2022iros}    & 54,8                        & 8.8                       & n/a                             \\
		\fdmos, delayed~\cite{mersch2022ral} & 75.5                        & 70.9                      & \textbf{44.8}                   \\
		\fdmos, online                       & 71.1                        & 68.7                      & 34.6                            \\
		Ours, \oursscan                      & 77.0                        & 79.2                      & 36.8                            \\
		Ours, \oursdelay                     & \textbf{78.4}               & \textbf{81.7}             & 40.3                            \\
		\bottomrule
	\end{tabular}
	\caption{Generalization capabilities of different methods on datasets outside of the training distribution. We report the average moving~\ac*{iou}. Best results in bold.}
	\label{tab:generalization}
	\vspace{-0.3cm}
\end{table}

\subsection{\Belief}
\label{exp:ablation}

Next, we carry out experiments that show how our proposed \belief can improve moving~\ac{iou}, recall, and precision. We compare the prediction of our model for the current scan (referred to as ``\oursscan'') to our \belief after fusing only the scan prediction (referred to as ``\oursbelief'').

One can see from~\tabref{tab:ablation} that the probabilistic fusion using a \bayesfilter consistently increases the precision of our scan prediction by rejecting false positives in previously predicted regions. At the same time, the recall drops due to the discretization error between ground points and the boundary of moving objects. Next, we additionally fuse the \map points that we predict to be moving (referred to as ``\oursmap''). The results indicate that additionally fusing the \map predictions increases the recall compared to the \belief that only integrates scan predictions.

Our last setup (referred to as ``\oursdelay'') first integrates~$10$ scan and moving \map predictions into our \belief before querying it for evaluation as explained in~\secref{exp:kitti}. This setup again achieves the best result in terms of IoU on most of the sequences since we can now use the \map predictions to identify traces of moving objects and update the \belief accordingly, even if the previous scan-based prediction was static. Solely in the case of \apollo, the setup using the \belief only fusing scan predictions is slightly better in terms of moving~\ac*{iou}. Since the recall of moving objects in the scan predictions is already very high for \apollo, we believe that the negative impact of discretization errors from additionally fusing moving \map points is more dominant in the final~\ac*{iou} than the improvement from correcting false negatives.

\begin{table*}[]
	\centering
	\begin{tabular}{l|ccc|ccc|ccc|ccc}
		\toprule
		            & \multicolumn{3}{c}{\semkitti~\cite{behley2019iccv,chen2021ral}} & \multicolumn{3}{c}{KITTI~\cite{geiger2012cvpr}} & \multicolumn{3}{c}{\apollo~\cite{lu2019cvpr}} & \multicolumn{3}{c}{\nuscenes~\cite{caesar2020cvpr}}                                                                                                                                 \\
		            & \multicolumn{3}{c}{Validation 08}                               & \multicolumn{3}{c}{Tracking 19}                 & \multicolumn{3}{c}{}                          & \multicolumn{3}{c}{Validation}                                                                                                                                                      \\
		Method      & IoU                                                             & R                                               & P                                             & IoU                                                 & R             & P             & IoU           & R             & P             & IoU           & R             & P             \\
		\midrule
		\oursscan   & 83.8                                                            & 87.5                                            & 95.3                                          & 77.0                                                & \textbf{84.6} & 89.6          & 79.2          & \textbf{93.0} & 84.5          & 36.8          & 43.4          & 70.0          \\
		\oursbelief & 84.0                                                            & 86.4                                            & \textbf{96.8}                                 & 76.7                                                & 80.3          & \textbf{94.4} & \textbf{82.1} & 92.3          & \textbf{88.6} & 36.6          & 40.8          & \textbf{81.1} \\
		\oursmap    & 83.9                                                            & 86.7                                            & 96.3                                          & 76.9                                                & 81.8          & 92.8          & 81.3          & 92.4          & 87.7          & 36.9          & 41.7          & 79.4          \\
		\oursdelay  & \textbf{86.1}                                                   & \textbf{88.7}                                   & \textbf{96.8}                                 & \textbf{78.4}                                       & 83.4          & 92.9          & 81.7          & 92.9          & 87.7          & \textbf{40.3} & \textbf{45.7} & 77.9          \\
		\bottomrule
	\end{tabular}
	\caption{Ablation study on average moving~\ac*{iou}, recall (R), and precision (P) in \% for our scan-based prediction and different \belief fusion strategies.}
	\label{tab:ablation}
	\vspace{-0.3cm}
\end{table*}

\subsection{Online Mapping}
\label{exp:mapping}
\begin{figure*}[t]
	\centering
	\begingroup%
  \makeatletter%
  \providecommand\color[2][]{%
    \errmessage{(Inkscape) Color is used for the text in Inkscape, but the package 'color.sty' is not loaded}%
    \renewcommand\color[2][]{}%
  }%
  \providecommand\transparent[1]{%
    \errmessage{(Inkscape) Transparency is used (non-zero) for the text in Inkscape, but the package 'transparent.sty' is not loaded}%
    \renewcommand\transparent[1]{}%
  }%
  \providecommand\rotatebox[2]{#2}%
  \newcommand*\fsize{\dimexpr\f@size pt\relax}%
  \newcommand*\lineheight[1]{\fontsize{\fsize}{#1\fsize}\selectfont}%
  \ifx\svgwidth\undefined%
    \setlength{\unitlength}{462.76698303bp}%
    \ifx\svgscale\undefined%
      \relax%
    \else%
      \setlength{\unitlength}{\unitlength * \real{\svgscale}}%
    \fi%
  \else%
    \setlength{\unitlength}{\svgwidth}%
  \fi%
  \global\let\svgwidth\undefined%
  \global\let\svgscale\undefined%
  \makeatother%
  \begin{picture}(1,0.59650417)%
    \lineheight{1}%
    \setlength\tabcolsep{0pt}%
    \put(0,0){\includegraphics[width=\unitlength,page=1]{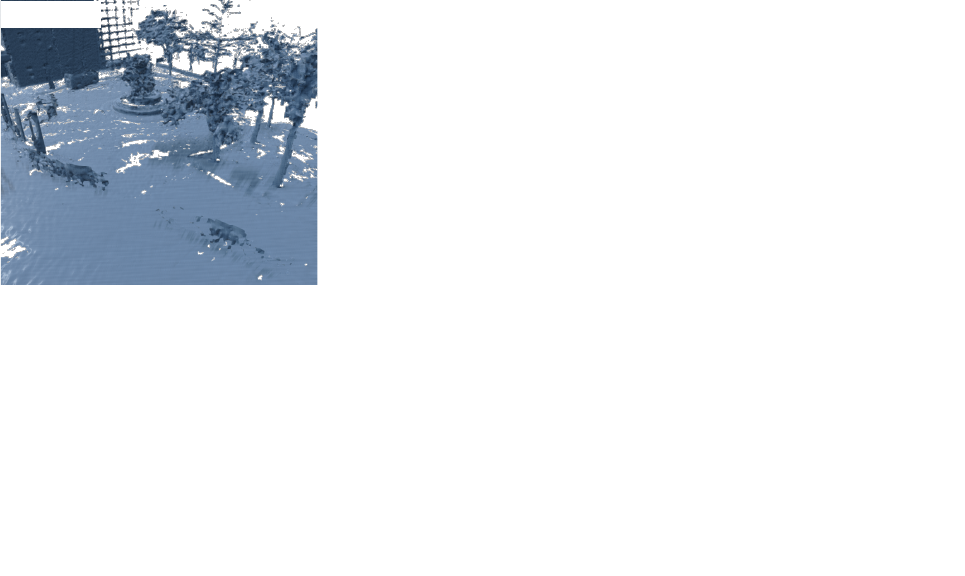}}%
    \put(0.03596321,0.57549382){\color[rgb]{0,0,0}\makebox(0,0)[lt]{\lineheight{1.25}\smash{\begin{tabular}[t]{l}Raw\end{tabular}}}}%
    \put(0,0){\includegraphics[width=\unitlength,page=2]{mapping.pdf}}%
    \put(0.35170944,0.5756274){\color[rgb]{0,0,0}\makebox(0,0)[lt]{\lineheight{1.25}\smash{\begin{tabular}[t]{l}4DMOS\end{tabular}}}}%
    \put(0,0){\includegraphics[width=\unitlength,page=3]{mapping.pdf}}%
    \put(0.70424379,0.57549377){\color[rgb]{0,0,0}\makebox(0,0)[lt]{\lineheight{1.25}\smash{\begin{tabular}[t]{l}Ours\end{tabular}}}}%
    \put(0,0){\includegraphics[width=\unitlength,page=4]{mapping.pdf}}%
  \end{picture}%
\endgroup%

	\caption{Reconstructed surfaces obtained with \vdbfusion~\cite{vizzo2022sensors} for CYT\_02~\cite{lin2019iros-larl} using a Livox MID40 scanner in the top row and \kittitracking~\cite{geiger2012cvpr} sequence 19 in the bottom row. \textit{Left:} Integrating both moving and non-moving points. \textit{Middle:} Integrating static points based on \fdmos~\cite{mersch2022ral} predictions using the \strategy. \textit{Right:} Integrating static points using our \belief after waiting~$10$ frames. Solid markers show remaining traces from moving objects in the map and the dashed marker indicates a static area that 4DMOS removed due to false positive predictions.}
	\label{fig:mapping}
	\vspace{-0.3cm}
\end{figure*}

Finally, we analyze how we can use our approach and the corresponding \belief for online mapping. We use the \vdbfusion~\cite{vizzo2022sensors} library that provides a TSDF-based reconstruction pipeline using the VDB data structure to build a final 3D model~\cite{museth2013siggraph}. We show the results in~\figref{fig:mapping} for the CYT\_02 sequence~\cite{lin2019iros-larl} (top row) and for the \kittitracking sequence 19 (bottom row). The CYT\_02 data was obtained with a Livox MID40 scanner, which has a smaller field of view and an irregular sampling pattern compared to rotating 3D LiDARs. This makes it harder to identify moving objects from a limited sequence of frames.

The first column shows the reconstructed surfaces from integrating all scans, including moving points. One can see the traces of moving objects in the map which are undesirable for planning.

The middle column shows the reconstruction after integrating the static predictions from \fdmos using the \strategy. Although \fdmos removes most of the dynamic traces, some moving objects remain in the map, as indicated by the solid markers in~\figref{fig:mapping}. When used with the Livox scanner, \fdmos removes a lot of static points due to the irregular sampling pattern and the limited number of past scans as encircled by the dashed marker in~\figref{fig:mapping}. We show our final map in the right column.

Based on the high recall achieved in~\secref{exp:ablation}, we query the \belief after fusing~$10$ scan and \map predictions. To additionally handle the discretization error close to the ground as explained in~\secref{sec:mapping}, we only integrate points for which both the map belief and the corresponding scan prediction are static. By doing so, we can achieve sub-voxel accuracy and even remove moving points that are close to the ground.

\section{Conclusion}
\label{sec:conclusion}
In this paper, we presented a novel approach to segment moving objects in the current scan and \map. We use a sparse 4D CNN to jointly extract spatio-temporal features based on the discrepancy between scan and map as well as the relative timestamps between points. Additionally, we suggest fusing our predictions into a probabilistic \belief. This allows us to successfully segment moving objects and even recover from false positive predictions. We evaluated our approach on different datasets with different sensor setups and demonstrated its effectiveness and generalization capabilities. Finally, we carried out experiments to evaluate the impact of our \belief and show that it improves the precision and recall of our~\ac*{mos} and can be effectively used to construct a static representation of the environment online.

\section*{Acknowledgments}
We thank Hyungtae Lim for fruitful discussions and Jiadai Sun for providing baseline results.

\bibliographystyle{plain_abbrv}
\bibliography{mersch2023ral}

\begin{thebibliography}{10}

\bibitem{arora2021ecmr}
M.~Arora, L.~Wiesmann, X.~Chen, and C.~Stachniss.
\newblock {Mapping the Static Parts of Dynamic Scenes from 3D LiDAR Point
  Clouds Exploiting Ground Segmentation}.
\newblock In {\em Proc.~of the Europ.~Conf.~on Mobile Robotics (ECMR)}, 2021.

\bibitem{arora2023jras}
M.~Arora, L.~Wiesmann, X.~Chen, and C.~Stachniss.
\newblock {Static Map Generation from 3D LiDAR Point Clouds Exploiting Ground
  Segmentation}.
\newblock {\em Journal on Robotics and Autonomous Systems (RAS)}, 159:104287,
  2023.

\bibitem{behley2021ijrr}
J.~Behley, M.~Garbade, A.~Milioto, J.~Quenzel, S.~Behnke, J.~Gall, and
  C.~Stachniss.
\newblock {Towards 3D LiDAR-based Semantic Scene Understanding of 3D Point
  Cloud Sequences: The SemanticKITTI Dataset}.
\newblock {\em Intl.~Journal~of Robotics Research (IJRR)}, 40(8--9):959--967,
  2021.

\bibitem{behley2019iccv}
J.~Behley, M.~Garbade, A.~Milioto, J.~Quenzel, S.~Behnke, C.~Stachniss, and
  J.~Gall.
\newblock {SemanticKITTI: A Dataset for Semantic Scene Understanding of LiDAR
  Sequences}.
\newblock In {\em Proc.~of the IEEE/CVF Intl.~Conf.~on Computer Vision (ICCV)},
  2019.

\bibitem{biber2005rss}
P.~Biber and T.~Duckett.
\newblock {Dynamic Maps for Long-Term Operation of Mobile Service Robots}.
\newblock In {\em Proc.~of Robotics: Science and Systems (RSS)}, 2005.

\bibitem{caesar2020cvpr}
H.~Caesar, V.~Bankiti, A.~Lang, S.~Vora, V.~Liong, Q.~Xu, A.~Krishnan, Y.~Pan,
  G.~Baldan, and O.~Beijbom.
\newblock {nuScenes: A Multimodal Dataset for Autonomous Driving}.
\newblock In {\em Proc.~of the IEEE/CVF Conf.~on Computer Vision and Pattern
  Recognition (CVPR)}, 2020.

\bibitem{chen2021ral}
X.~Chen, S.~Li, B.~Mersch, L.~Wiesmann, J.~Gall, J.~Behley, and C.~Stachniss.
\newblock {Moving Object Segmentation in 3D LiDAR Data: A Learning-based
  Approach Exploiting Sequential Data}.
\newblock {\em IEEE Robotics and Automation Letters (RA-L)}, 6(4):6529--6536,
  2021.

\bibitem{chen2022ral}
X.~Chen, B.~Mersch, L.~Nunes, R.~Marcuzzi, I.~Vizzo, J.~Behley, and
  C.~Stachniss.
\newblock {Automatic Labeling to Generate Training Data for Online LiDAR-Based
  Moving Object Segmentation}.
\newblock {\em IEEE Robotics and Automation Letters (RA-L)}, 7(3):6107--6114,
  2022.

\bibitem{chen2019iros}
X.~Chen, A.~Milioto, E.~Palazzolo, P.~Giguère, J.~Behley, and C.~Stachniss.
\newblock {SuMa++: Efficient LiDAR-based Semantic SLAM}.
\newblock In {\em Proc.~of the IEEE/RSJ Intl.~Conf.~on Intelligent Robots and
  Systems (IROS)}, 2019.

\bibitem{choy2019cvpr}
C.~Choy, J.~Gwak, and S.~Savarese.
\newblock {4D Spatio-Temporal ConvNets: Minkowski Convolutional Neural
  Networks}.
\newblock In {\em Proc.~of the IEEE/CVF Conf.~on Computer Vision and Pattern
  Recognition (CVPR)}, 2019.

\bibitem{everingham2010ijcv}
M.~Everingham, L.~Van~Gool, C.~Williams, J.~Winn, and A.~Zisserman.
\newblock {The Pascal Visual Object Classes (VOC) Challenge}.
\newblock {\em Intl.~Journal~of Computer Vision (IJCV)}, 88(2):303--338, 2010.

\bibitem{gehrung2017isprsannals}
J.~Gehrung, M.~Hebel, M.~Arens, and U.~Stilla.
\newblock {An Approach to Extract Moving Objects From MLS Data Using a
  Volumetric Background Representation}.
\newblock {\em ISPRS Annals of the Photogrammetry, Remote Sensing and Spatial
  Information Sciences}, IV-1/W1:107--114, 2017.

\bibitem{geiger2012cvpr}
A.~Geiger, P.~Lenz, and R.~Urtasun.
\newblock {Are we ready for Autonomous Driving? The KITTI Vision Benchmark
  Suite}.
\newblock In {\em Proc.~of the IEEE Conf.~on Computer Vision and Pattern
  Recognition (CVPR)}, 2012.

\bibitem{henein2018arxiv}
M.~Henein, G.~Kennedy, V.~Ila, and R.~Mahony.
\newblock {Simultaneous Localization and Mapping with Dynamic Rigid Objects}.
\newblock {\em arXiv preprint}, arXiv:1805.03800, 2018.

\bibitem{hoermann2018icra}
S.~Hoermann, M.~Bach, and K.~Dietmayer.
\newblock Dynamic occupancy grid prediction for urban autonomous driving: A
  deep learning approach with fully automatic labeling.
\newblock In {\em Proc.~of the IEEE Intl.~Conf.~on Robotics \& Automation
  (ICRA)}, 2018.

\bibitem{hornung2013ar}
A.~Hornung, K.~Wurm, M.~Bennewitz, C.~Stachniss, and W.~Burgard.
\newblock {OctoMap: An Efficient Probabilistic 3D Mapping Framework Based on
  Octrees}.
\newblock {\em Autonomous Robots}, 34(3):189--206, 2013.

\bibitem{huang2022eccv}
S.~Huang, Z.~Gojcic, J.~Huang, A.~Wieser, and K.~Schindler.
\newblock {Dynamic 3D Scene Analysis by Point Cloud Accumulation}.
\newblock In {\em Proc.~of the Europ.~Conf.~on Computer Vision (ECCV)}, 2022.

\bibitem{kim2020iros}
G.~Kim and A.~Kim.
\newblock {Remove, Then Revert: Static Point Cloud Map Construction Using
  Multiresolution Range Images}.
\newblock In {\em Proc.~of the IEEE/RSJ Intl.~Conf.~on Intelligent Robots and
  Systems (IROS)}, 2020.

\bibitem{kim2022ral}
J.~Kim, J.~Woo, and Sunghoon.
\newblock {RVMOS: Range-View Moving Object Segmentation Leveraged by Semantic
  and Motion Features}.
\newblock {\em IEEE Robotics and Automation Letters (RA-L)}, 7(3):8044--8051,
  2022.

\bibitem{kreutz2023wacv}
T.~Kreutz, M.~M\"uhlh\"auser, and A.S. Guinea.
\newblock {Unsupervised 4D LiDAR Moving Object Segmentation in Stationary
  Settings with Multivariate Occupancy Time Series}.
\newblock In {\em Proc.~of the IEEE Winter Conf.~on Applications of Computer
  Vision (WACV)}, 2023.

\bibitem{kuemmerle2013icra}
R.~K\"ummerle, M.~Ruhnke, B.~Steder, C.~Stachniss, and W.~Burgard.
\newblock {A Navigation System for Robots Operating in Crowded Urban
  Environments}.
\newblock In {\em Proc.~of the IEEE Intl.~Conf.~on Robotics \& Automation
  (ICRA)}, 2013.

\bibitem{lim2021ral}
H.~Lim, S.~Hwang, and H.~Myung.
\newblock {ERASOR: Egocentric Ratio of Pseudo Occupancy-Based Dynamic Object
  Removal for Static 3D Point Cloud Map Building}.
\newblock {\em IEEE Robotics and Automation Letters (RA-L)}, 6(2):2272--2279,
  2021.

\bibitem{lin2019iros-larl}
J.~Lin and F.~Zhang.
\newblock {Loam\_livox A Robust LiDAR Odemetry and Mapping LOAM Package for
  Livox LiDAR}.
\newblock In {\em Proc.~of the IEEE/RSJ Intl.~Conf.~on Intelligent Robots and
  Systems (IROS)}, 2019.

\bibitem{lu2019cvpr}
W.~Lu, Y.~Zhou, G.~Wan, S.~Hou, and S.~Song.
\newblock {L3-Net: Towards Learning Based LiDAR Localization for Autonomous
  Driving}.
\newblock In {\em Proc.~of the IEEE/CVF Conf.~on Computer Vision and Pattern
  Recognition (CVPR)}, 2019.

\bibitem{mersch2022ral}
B.~Mersch, X.~Chen, I.~Vizzo, L.~Nunes, J.~Behley, and C.~Stachniss.
\newblock {Receding Moving Object Segmentation in 3D LiDAR Data Using Sparse 4D
  Convolutions}.
\newblock {\em IEEE Robotics and Automation Letters (RA-L)}, 7(3):7503--7510,
  2022.

\bibitem{milioto2019iros}
A.~Milioto, I.~Vizzo, J.~Behley, and C.~Stachniss.
\newblock {RangeNet++: Fast and Accurate LiDAR Semantic Segmentation}.
\newblock In {\em Proc.~of the IEEE/RSJ Intl.~Conf.~on Intelligent Robots and
  Systems (IROS)}, 2019.

\bibitem{museth2013siggraph}
K.~Museth, J.~Lait, J.~Johanson, J.~Budsberg, R.~Henderson, M.~Alden, P.~Cucka,
  D.~Hill, and A.~Pearce.
\newblock {OpenVDB: An Open-source Data Structure and Toolkit for
  High-resolution Volumes}.
\newblock In {\em ACM SIGGRAPH 2013 courses}. 2013.

\bibitem{niessner2013siggraph}
M.~Nie{\ss}ner, M.~Zollh{\"o}fer, S.~Izadi, and M.~Stamminger.
\newblock {Real-time 3D Reconstruction at Scale using Voxel Hashing}.
\newblock In {\em Proc.~of the SIGGRAPH Asia}, 2013.

\bibitem{nuss2018ijrr}
D.~Nuss, S.~Reuter, M.~Thom, T.~Yuan, G.~Krehl, M.~Maile, A.~Gern, and
  K.~Dietmayer.
\newblock {A Random Finite Set Approach for Dynamic Occupancy Grid Maps with
  Real-Time Application}.
\newblock {\em Intl.~Journal~of Robotics Research (IJRR)}, 37(8):841--866,
  2018.

\bibitem{pagad2020icra}
S.~Pagad, D.~Agarwal, S.~Narayanan, K.~Rangan, H.~Kim, and G.~Yalla.
\newblock {Robust Method for Removing Dynamic Objects from Point Clouds}.
\newblock In {\em Proc.~of the IEEE Intl.~Conf.~on Robotics \& Automation
  (ICRA)}, 2020.

\bibitem{pfreundschuh2021icra}
P.~Pfreundschuh, H.F.C. Hendrikx, V.~Reijgwart, R.~Dub{\'e}, R.~Siegwart, and
  A.~Cramariuc.
\newblock {Dynamic Object Aware LiDAR SLAM based on Automatic Generation of
  Training Data}.
\newblock In {\em Proc.~of the IEEE Intl.~Conf.~on Robotics \& Automation
  (ICRA)}, 2021.

\bibitem{pomerleau2014icra}
F.~Pomerleau, P.~Kr{\"u}siand, F.~Colas, P.~Furgale, and R.~Siegwart.
\newblock {Long-term 3D Map Maintenance in Dynamic Environments}.
\newblock In {\em Proc.~of the IEEE Intl.~Conf.~on Robotics \& Automation
  (ICRA)}, 2014.

\bibitem{ruchti2018icra}
P.~Ruchti and W.~Burgard.
\newblock {Mapping with Dynamic-Object Probabilities Calculated from Single 3D
  Range Scans}.
\newblock In {\em Proc.~of the IEEE Intl.~Conf.~on Robotics \& Automation
  (ICRA)}, 2018.

\bibitem{schauer2018ral}
J.~Schauer and A.~N{\"u}chter.
\newblock {The Peopleremover -- Removing Dynamic Objects From 3-D Point Cloud
  Data by Traversing a Voxel Occupancy Grid}.
\newblock {\em IEEE Robotics and Automation Letters (RA-L)}, 3(3):1679--1686,
  2018.

\bibitem{stachniss2005aaai}
C.~Stachniss and W.~Burgard.
\newblock {Mobile Robot Mapping and Localization in Non-Static Environments}.
\newblock In {\em Proc.~of the National Conf.~on Artificial Intelligence
  (AAAI)}, 2005.

\bibitem{sun2022iros}
J.~Sun, Y.~Dai, X.~Zhang, J.~Xu, R.~Ai, W.~Gu, and X.~Chen.
\newblock {Efficient Spatial-Temporal Information Fusion for LiDAR-Based 3D
  Moving Object Segmentation}.
\newblock In {\em Proc.~of the IEEE/RSJ Intl.~Conf.~on Intelligent Robots and
  Systems (IROS)}, 2022.

\bibitem{thrun2005probrobbook}
S.~Thrun, W.~Burgard, and D.~Fox.
\newblock {\em {Probabilistic Robotics}}.
\newblock MIT Press, 2005.

\bibitem{vizzo2022sensors}
I.~Vizzo, T.~Guadagnino, J.~Behley, and C.~Stachniss.
\newblock {VDBFusion: Flexible and Efficient TSDF Integration of Range Sensor
  Data}.
\newblock {\em Sensors}, 22(3), 2022.

\bibitem{vizzo2023ral}
I.~Vizzo, T.~Guadagnino, B.~Mersch, L.~Wiesmann, J.~Behley, and C.~Stachniss.
\newblock {KISS-ICP: In Defense of Point-to-Point ICP -- Simple, Accurate, and
  Robust Registration If Done the Right Way}.
\newblock {\em IEEE Robotics and Automation Letters (RA-L)}, 8(2):1029--1036,
  2023.

\bibitem{wellhausen2017ssrr}
L.~Wellhausen, R.~Dub{\'e}, A.~Gawel, R.~Siegwart, and C.~Cadena.
\newblock {Reliable Real-time Change Detection and Mapping for 3D LiDARs}.
\newblock In {\em Proc.~of the IEEE Intl.~Sym.~on Safety, Security, and Rescue
  Robotics (SSRR)}, 2017.

\bibitem{wurm2010icraws}
K.~Wurm, A.~Hornung, M.~Bennewitz, C.~Stachniss, and W.~Burgard.
\newblock {OctoMap: A Probabilistic, Flexible, and Compact 3D Map
  Representation for Robotic Systems}.
\newblock In {\em Workshop on Best Practice in 3D Perception and Modeling for
  Mobile Manipulation, IEEE Int. Conf. on Robotics \& Automation (ICRA)}, 2010.

\bibitem{yoon2019crv}
D.~Yoon, T.~Tang, and T.~Barfoot.
\newblock {Mapless Online Detection of Dynamic Objects in 3D Lidar}.
\newblock In {\em Proc.~of the Conf.~on Computer and Robot Vision (CRV)}, 2019.

\bibitem{zimmerman2022iros}
N.~Zimmerman, L.~Wiesmann, T.~Guadagnino, T.~Läbe, J.~Behley, and
  C.~Stachniss.
\newblock {Robust Onboard Localization in Changing Environments Exploiting Text
  Spotting}.
\newblock In {\em Proc.~of the IEEE/RSJ Intl.~Conf.~on Intelligent Robots and
  Systems (IROS)}, 2022.

\end{thebibliography}

\end{document}